\documentclass[runningheads]{llncs}

\usepackage{graphicx}
\usepackage{threeparttable}
\usepackage{float}
\usepackage{rotating}
\usepackage{xcolor}
\usepackage[table]{xcolor}
\usepackage{textcomp}
\usepackage{xcolor}
\usepackage{booktabs}
\usepackage{multirow}
\usepackage{makecell}
\usepackage{algorithm}
\usepackage{algpseudocode}
\usepackage{hyperref}
\usepackage{amsmath}
\usepackage{amssymb}
\usepackage{enumitem}
\usepackage{orcidlink}

\begin{document}

\title{Adaptive Multi-Granularity Temporal Modeling for Weakly Supervised Video Anomaly Detection}

\titlerunning{Adaptive Multi-Granularity Temporal Modeling for WSVAD}

\author{Changyi Li\inst{1}\thanks{Corresponding author.}\orcidlink{0009-0004-9034-9122} \and
Yu Xiao\inst{1}\orcidlink{0000-0002-4517-3779}}
\authorrunning{C. Li and Y. Xiao}
\institute{Aalto University, 02150 Espoo, Finland\\
\email{\{changyi.li, yu.xiao\}@aalto.fi}}

\maketitle

\begin{abstract}
As the scale of video surveillance data outpaces manual annotation capacities, weakly supervised video anomaly detection (WSVAD) has emerged as a critical research frontier. Most existing approaches formulate WSVAD within a Multiple Instance Learning (MIL) framework that relies on rigid, hand-crafted temporal priors to supervise anomaly scoring. However, such formulations exhibit limited adaptability to the wide variation in anomaly durations and temporal dynamics observed in real-world videos, often leading to unstable or unreliable snippet-level predictions.
To address this limitation, we propose an adaptive temporal modeling framework for WSVAD that explicitly accounts for variations in video dynamics across multiple temporal granularities. First, we introduce a Temporal Refinement Module (TRM) that leverages dynamic positional encoding and a learnable class token to model long-range temporal dependencies while distilling a stable global video-level representation. Second, to capture anomalous events with varying frequency and duration, we develop an adaptive Event Segmentation Module (ESM) that identifies event boundaries through temporal discontinuity analysis and aggregates snippet features into discriminative event-level representations. Finally, for snippet-level and event-level predictions, we propose an adaptive similarity-based fusion strategy that dynamically integrates anomaly scores into video-level predictions, replacing fixed \textit{top-k} aggregation heuristics with global semantic relevance. Extensive experiments on two benchmarks demonstrate that the proposed framework consistently outperforms state-of-the-art methods. 

\keywords{Video Anomaly Detection  \and Weakly Supervised Learning \and Event Segmentation.}
\end{abstract}

\section{Introduction}
Driven by the rapid proliferation of large-scale surveillance infrastructure, video anomaly detection (VAD) has become increasingly critical in a wide range of real-world applications, including industrial automation~\cite{liu2024ipad}, intelligent transportation~\cite{orlova2025simplifying}, and public security~\cite{sultani2018real}. The goal of VAD is to identify abnormal events or behaviors that deviate from established normal patterns in long, untrimmed videos. As surveillance systems scale in both volume and duration, relying on manual monitoring becomes increasingly impractical due to its labor-intensive nature, susceptibility to human error, and limited ability to provide continuous coverage. These challenges motivate the development of automated VAD systems that are both scalable and reliable.

To alleviate the prohibitive cost of fine-grained temporal annotation, weakly supervised video anomaly detection has emerged as a practical and widely adopted paradigm \cite{sultani2018real,pu2022locality,zhou2023dual,pu2024learning}. In this setting, videos are annotated only with coarse video-level labels indicating whether an anomaly occurs, without specifying its temporal location. Multiple Instance Learning (MIL)~\cite{dietterich1997solving} naturally underpins most weakly supervised VAD approaches by treating each video as a bag of temporal snippets and inferring anomaly scores through snippet aggregation. Early works, such as Sultani et al.~\cite{sultani2018real}, formulate anomalous videos as positive bags and introduce ranking-based losses to distinguish them from normal ones. Subsequent studies enhance this framework by modeling temporal dynamics~\cite{pu2022locality} or incorporating memory mechanisms to improve robustness~\cite{zhou2023dual}. A more recent method further captures both long-term and short-term temporal dependencies for better performance in anomaly detection~\cite{pu2024learning}.

Despite notable progress, existing weakly supervised VAD methods share a fundamental limitation: they rely on fixed temporal assumptions that are poorly aligned with the diverse and variable temporal structures of real-world videos. For instance, Pu et al.~\cite{pu2024learning} employ a fixed-length window to compute short-term temporal dependencies. Additionally, under weak supervision, snippet-level predictions must be inferred implicitly, yet their reliability is often low—particularly during training stages when discriminative representations have not yet emerged~\cite{he2025self}. As a result, video-level optimization can be dominated by noisy or erroneous snippets. More critically, most MIL-based frameworks assume a canonical anomaly pattern and typically adopt fixed aggregation heuristics such as \textit{top-k} pooling~\cite{pu2022locality,zhou2023dual}. These assumptions significantly restrict adaptability to anomalies with diverse durations, frequencies, and temporal organizations.

We argue that addressing the inherent variability of video dynamics requires adaptive temporal modeling across multiple granularities, rather than relying on fixed heuristics. Motivated by this insight, we propose an adaptive weakly supervised VAD framework that explicitly models temporal structure at the snippet, event, and video levels.
First, we introduce a Temporal Refinement Module (TRM) to enable adaptive long-range temporal reasoning and stable global supervision. By prepending a learnable class token to the snippet sequence and updating it through self-attention, the TRM distills a robust video-level representation that captures global semantics independent of individual snippet predictions. To further enhance temporal expressiveness, we incorporate Dynamic Position Encoding (DPE), which adaptively models temporal ordering and long-range dependencies beyond fixed positional assumptions.

Second, to explicitly handle anomalies with highly variable temporal extents, we propose an Event Segmentation Module (ESM) that elevates modeling from snippets to events. The ESM identifies event boundaries by detecting temporal discontinuities in feature evolution and aggregates snippets within each segment into discriminative event-level representations. This adaptive event-centric modeling enables the framework to capture both short-lived and long-duration anomalies effectively.

Finally, instead of relying on fixed \textit{top-k} aggregation for video-level prediction, we introduce an adaptive similarity-based fusion strategy. By measuring semantic similarity between snippet- or event-level features and the global video representation, our method dynamically weights contributions from different temporal positions, providing a principled alternative to heuristic aggregation.

Extensive experiments on two benchmark datasets, UCF-Crime and XD-Violence, demonstrate that the proposed framework consistently outperforms state-of-the-art (SOTA) methods. Comprehensive ablation studies further validate the effectiveness of each component and highlight its complementary roles in addressing the core challenges of weakly supervised video anomaly detection.

The contributions of this work can be summarized as:
\begin{itemize}[label=$\bullet$]
    \item We propose an adaptive weakly supervised VAD framework that explicitly models temporal variability across snippet, event, and video levels.
    \item A Temporal Refinement Module with a class token and Dynamic Position Encoding is introduced to capture long-range temporal dependencies and stabilize video-level supervision.
    \item An Event Segmentation Module is introduced to adaptively capture anomalies with varying temporal durations via event-level representation learning.
    \item An adaptive similarity-based fusion strategy replaces fixed \textit{top-k} aggregation by dynamically weighting anomaly scores from different temporal positions.
\end{itemize}

\section{Proposed Method}
\label{sec:method}
In weakly supervised video anomaly detection (WSVAD), the training data consist of a set of videos annotated with video-level labels $Y \in \{0, 1\}$, where $Y=1$ denotes an anomalous video and $Y=0$ indicates a normal one. Each video is temporally segmented into $T$ non-overlapping segments. The objective is to train a model that, during inference, assigns an accurate anomaly score $s_t \in [0, 1]$ for each segment $t \in \{1, \dots, T\}$.

\begin{figure*}[t]
    \centering
    \includegraphics[width=\textwidth]{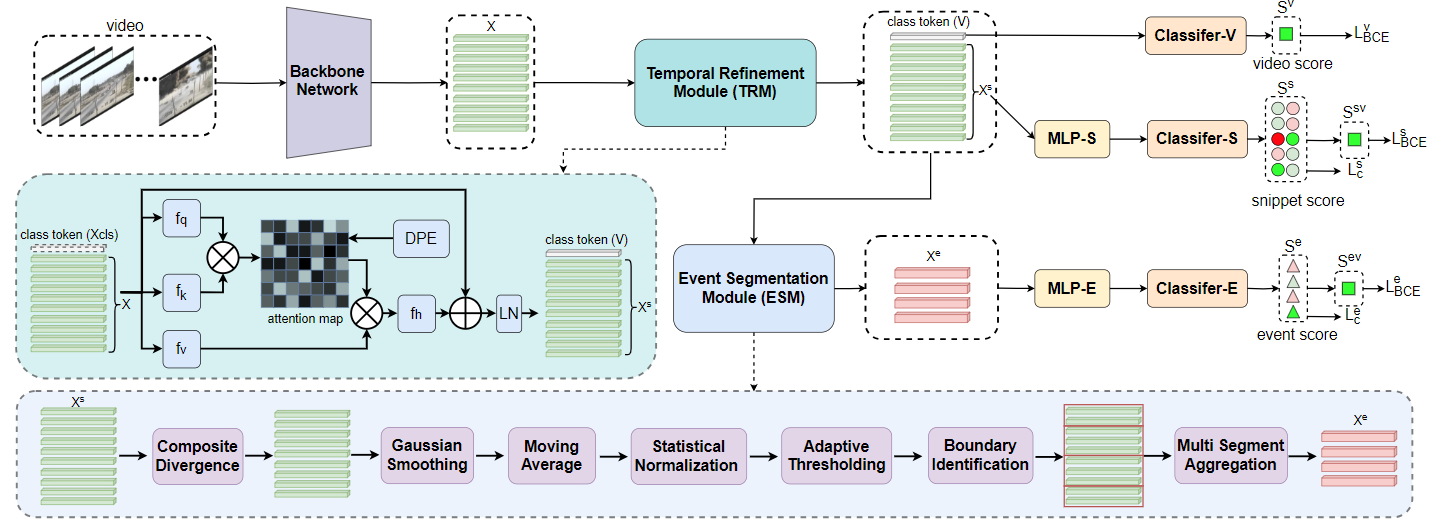}
    \caption{\textbf{Overview of the proposed framework.} Snippet-level features are first extracted from input videos using backbone networks. These features are then refined by the TRM to capture long-range temporal dependencies and produce a global video-level representation enriched with contextual information. Subsequently, the ESM is adaptively applied to identify event boundaries and aggregate snippet features into event-level representations. Finally, the model is jointly optimized through a multi-scale prediction strategy that integrates snippet-, event-, and video-level outputs.
    }
    \label{fig:framework}
\end{figure*}

The overall architecture of the proposed framework is illustrated in Fig.~\ref{fig:framework}. Given an untrimmed video, it is first divided into a sequence of non-overlapping snippets using a sliding window of 16 frames. A pre-trained neural network (I3D backbone~\cite{carreira2017quo}) is then employed to extract spatio-temporal features for each snippet. These features are concatenated along the temporal dimension to form a video-level feature representation $X \in \mathbb{R}^{T \times D}$, where $T$ denotes the number of snippets and $D$ represents the feature dimensionality. 
To enhance temporal representation, a Temporal Refinement Module (TRM) is applied to model temporal dependencies among snippets. By introducing a learnable class token, the TRM aggregates sequence-level information and produces a global video representation $V$, while simultaneously generating refined snippet features $X^s$.
These refined features are subsequently fed into the Event Segmentation Module (ESM) to detect precise temporal boundaries, enabling the extraction of event-level features $X^e$.
Both snippet-level and event-level features are then projected into robust semantic embedding spaces via two-layer multilayer perceptrons (MLPs). Finally, three causal convolutional layers serve as classifiers to predict anomaly scores at the video $S^{v}$, snippet $S^{s}$, and event $S^{e}$ levels. The event-level and snippet-level scores are further integrated to produce the final video-level anomaly predictions, denoted as $S^{ev}$ and $S^{sv}$, respectively.

\subsection{Temporal Refinement Module (TRM)}

As shown in Fig.~\ref{fig:framework}, the Temporal Refinement Module (TRM) is designed to convert independent snippet-level features into context-aware representations by explicitly modeling long-range temporal dependencies. Given a sequence of snippet features $X = \{x_i\}_{i=1}^T$ extracted by the I3D backbone~\cite{carreira2017quo}, a learnable class token $x_{cls} \in \mathbb{R}^D$ is prepended to the sequence to enable global context aggregation. This yields an augmented input $X_0 = [x_{cls}; x_1; x_2; \dots; x_T] \in \mathbb{R}^{(T+1) \times D}$, where the class token serves as a global representation that captures information across the entire temporal span of the video.

The TRM refines the input representations using a modified self-attention mechanism combined with residual learning. The augmented sequence $X_0$ is first projected into query, key, and value spaces via linear transformations $f_q(\cdot)$, $f_k(\cdot)$, and $f_v(\cdot)$, respectively. The attention affinity matrix $M$ is computed as
\begin{equation}
    M = f_q(X_0) \cdot f_k(X_0)^\top,
\end{equation}

Standard self-attention mechanisms are permutation-invariant and therefore lack explicit awareness of temporal order. To address this limitation, Dynamic Position Encoding (DPE), denoted by $PE$, is introduced, which injects localized temporal priors directly into the attention computation. Unlike conventional fixed positional encoding based on sinusoidal functions~\cite{vaswani2017attention}, the proposed DPE adapts naturally to varying video lengths through learnable parameters. Specially, for position $i$ and $j$, the positional bias $PE_{i,j}$ is defined as 
\begin{equation}
PE_{i,j} = \exp\left(-\left|\gamma(i - j)^\alpha - \beta\right|\right),
\end{equation}
where $\gamma$ and $\beta$ are learnable scalars. $\alpha$ is the hyperparameter that controls the temporal decay rate. This Gaussian-like formulation suppresses the influence of temporally distant snippets by assigning them lower attention weights, thereby emphasizing local temporal relationships while remaining sensitive to non-linear temporal patterns. Therefore, the gated attention map $A$ is computed as:
\begin{equation}
    A = \text{softmax}\left(\frac{M}{\sqrt{D_h}} + PE\right),
\end{equation}
Where $D_h$ denotes the hidden dimension. 

The context-aware representations $X^c$ are then obtained by aggregating the value features:
\begin{equation}
    [V, X^c] = A \cdot f_v(X_0), 
\end{equation}
where $V$ corresponds to the output associated with the class token and represents a refined global video embedding learned during training. To preserve the original spatio-temporal information and ensure stable gradient propagation, a residual connection followed by layer normalization is applied. The final refined snippet representations $X^s$ are given by
\begin{equation}
    X^s = \mathrm{LN}\left(X + f_h\left(X^c\right)\right),
\end{equation}
where $\mathrm{LN}(\cdot)$ denotes layer normalization and $f_h(\cdot)$ is a projection head used to align feature dimensions. The TRM effectively captures global temporal context while maintaining fine-grained local cues from the backbone features.

\subsection{Event Segmentation Module (ESM)}
While snippet-level features refined by the TRM provide fine-grained temporal cues, individual snippets lack sufficient temporal context to represent complete anomalous events. Moreover, anomalies vary substantially in both frequency and duration across different videos, rendering fixed-length temporal modeling inadequate. To address these limitations, we introduce the Event Segmentation Module (ESM), which aims to adaptively identify event boundaries and aggregate temporally coherent snippets into event-level representations. By explicitly modeling variable-length events, the ESM captures semantically consistent structures that are more informative than isolated snippets while remaining more precise than a single global video representation.

As illustrated in Fig.~\ref{fig:framework}, the ESM operates on the refined snippet features $X^s = \{x^s_t\}_{t=1}^T$ produced by the TRM. The ESM first computes a composite divergence between consecutive snippet features to characterize semantic transitions along the temporal dimension:
\begin{equation}
D_t = |x^s_t - x^s_{t+1}|^2 + \big(1 - \mathrm{cos}(x^s_t, x^s_{t+1})\big),
\end{equation}
where $D_t$ captures variations in both feature magnitude and directional orientation. 
To suppress high-frequency noise and stabilize the divergence signal, a Gaussian smoothing kernel is applied to the divergence sequence $\{D_t\}$, yielding the smoothed divergence $\tilde{D}_t$.
Next, a local moving average is computed over a sliding temporal window of size $W$:
\begin{equation}
MA_t = \frac{1}{W} \sum_{k=t-W/2}^{t+W/2} \tilde{D}_k,
\end{equation}
Using this local background estimate, the smoothed divergence is statistically normalized by forming a signal ratio:
\begin{equation}
SR_t = \frac{\tilde{D}_t}{MA_t},
\end{equation}
which highlights salient semantic transitions relative to the surrounding temporal context.

Event boundaries are then identified through adaptive thresholding based on the Median Absolute Deviation (MAD)~\cite{shao2025eventvad}:
\begin{equation}
\mathrm{Threshold} = \mathrm{median}(SR_t) + \lambda \cdot \mathrm{MAD}(SR_t),
\end{equation}
where $\lambda$ is a sensitivity hyperparameter. A temporal boundary is identified whenever $SR_t$ exceeds the threshold, resulting in a set of semantically coherent events:
\begin{equation}
E = \{(t_{k,s},t_{k,e})\}_{k=1}^K, 
\end{equation}
where $t_{k,s}$ and $t_{k,e}$ denote the start and end indices of the $k$-th event, respectively. 
Finally, for each detected event, the snippet features within its temporal span are aggregated to form an event-level representation. Specifically, the $k$-th event feature $x^e_k$ is obtained via weighted aggregation:
\begin{equation}
    x^e_k = \sum_{i=t_{k,s}}^{t_{k,e}} w_i \cdot x^s_i,
\end{equation}
where $x^s_i$ denotes the refined feature of the $i$-th snippet. The weights $w_i$ can be implemented either as a uniform average, $w_i = 1/(t_{k,e} - t_{k,s} + 1)$, or derived from attention scores produced by the TRM. This aggregation ensures that each event-level representation captures the dominant semantic dynamics of its corresponding temporal interval.

\subsection{Adaptive Score Fuse}
During training, the global video representation $V$, which encapsulates holistic temporal semantics, is directly input into its classifier to generate a video-level anomaly score. In contrast, the refined snippet-level features $X^s$ and the aggregated event-level features $X^e$ are independently processed by a two-layer Multi-Layer Perceptron (MLP) equipped with GELU activations and dropout to enhance non-linear representation capability. The resulting embeddings are subsequently fed into $1 \times 1$ causal convolutional layers and a sigmoid function, which serve as classifiers to produce snippet-level and event-level anomaly scores. Formally, the process is defined as:
\begin{align}
    S^v &= \text{Classifier-V}(V),\\
    S^s &= \text{Classifier-S}(\text{MLP-S}(X^s)), \\
    S^e &= \text{Classifier-E}(\text{MLP-E}(X^e)),
\end{align}
where $S^v \in \mathbb{R}$, $S^s \in \mathbb{R}^{1 \times T}$, and $S^{e} \in \mathbb{R}^{1 \times K}$ denote the predicted anomaly scores at the video, snippet, and event levels, respectively.

Since only video-level ground truth annotations are available during training, the snippet-level and event-level scores must be aggregated into video-level predictions to enable loss computation and optimization. 
Existing methods~\cite{pu2022locality,pu2024learning} typically adopt a top-$k$ pooling strategy, where the average of the highest $k$ scores is used as the video-level prediction. However, the effectiveness of such approaches is highly sensitive to the choice of $k$. Given that anomalous events exhibit significant variability in frequency and duration across videos—and even within the same scene—fixed aggregation strategies often lack robustness and generalization capability. 

To overcome this limitation, we propose an adaptive fusion mechanism that derives video-level anomaly scores by leveraging semantic consistency with the global video representation $V$. Specifically, we compute similarity-based weights between individual features and the global representation to dynamically modulate their contributions. Taking snippet-level fusion as an example, the video-level score is computed as
\begin{equation}
w_i = \mathrm{Softmax}\big(\mathrm{sim}(x^{s}_i, V)\big), \quad
S^{sv} = \sum_{i} w_i \cdot S^{s}_i,
\end{equation}
where $\text{sim}(\cdot, \cdot)$ denotes a similarity function such as cosine similarity, $x^s_i$ represents the $i$-th snippet feature, and $S^s_i$ is its corresponding anomaly score. By using the global video embedding as a reference, this adaptive weighting strategy enables the model to emphasize semantically relevant temporal segments while suppressing irrelevant noise, resulting in more robust video-level predictions. The same fusion process is applied to event-level scores to get $S^{ev}$.

Following standard practice~\cite{pu2022locality}, we adopt Binary Cross-Entropy (BCE) as the primary classification loss. For a mini-batch of size $B$, the loss is defined as
\begin{equation}
\mathcal{L}_{\mathrm{BCE}} = -\frac{1}{B} \sum_{i=1}^{B}
\big[ y_i \log(S_i) + (1 - y_i) \log(1 - S_i) \big],
\end{equation}
where $S_i$ denotes the predicted video-level anomaly score and $y_i$ is the corresponding ground-truth label. This loss is applied to video-level predictions obtained from snippet-level, event-level, and global representations, thereby providing supervision across multiple temporal granularities.

To further suppress false alarms in normal videos, a center loss $L_c$ is introduced to encourage temporal smoothness by penalizing large deviations among segment-level scores:
\begin{equation}
\mathcal{L}_{c} =
\mathbb{I}(Y = 0)\,\frac{1}{T}\sum_{t=1}^{T}(s_t - \bar{s})^2,
\end{equation}
where
$\bar{s} = \frac{1}{T}\sum_{t=1}^{T}s_t$ denotes the temporal mean score, and $\mathbb{I}(\cdot)$ is the indicator function.
To optimize our model across different temporal granularities, the loss functions at the video ($L_v$), snippet ($L_s$), and event ($L_e$) levels are defined as follows: 
\begin{align}
    \mathcal{L}_{v} &= \mathcal{L}^v_{BCE},\\
    \mathcal{L}_{s} &= \mathcal{L}^s_{BCE} + \lambda_s \mathcal{L}^s_{c}, \\
    \mathcal{L}_{e} &= \mathcal{L}^e_{BCE} + \lambda_e \mathcal{L}^e_{c},
\end{align}
where $\lambda_s$ and $\lambda_e$ are hyperparameters that balance the contribution of each loss term. The overall training objective, $L_{total}$, is defined as:
\begin{equation}
\mathcal{L}_{\mathrm{total}} =
\mathcal{L}_{s} +
\lambda_1 \mathcal{L}_{e} +
\lambda_2 \mathcal{L}_{v},
\end{equation}
where $\lambda_1$ and $\lambda_2$ are hyperparameters that trade off the respective loss terms. During testing, following~\cite{pu2024learning}, a score smoothing strategy is introduced to mitigate the impact of transient noise, and the snippet-level scores $S^s$ are used as the final predictions.

\section{Experiment}
\subsection{Datasets and Evaluation Metrics}

\textbf{UCF-Crime} is a large-scale real-world dataset for video anomaly detection, comprising 1,900 long, untrimmed videos. The dataset covers 13 types of anomalous events (e.g., Abuse, Arrest, Arson, and Fighting), along with a set of normal videos. Following the standard protocol, the training set consists of 1,610 videos with video-level labels, while the test set contains 290 videos with frame-level ground-truth annotations. 

\textbf{XD-Violence} is currently the largest multi-scene dataset for violence detection, encompassing 4,754 videos. It includes six categories of violent events collected from both real-world surveillance footage and movie scenes. Compared with UCF-Crime, XD-Violence poses greater challenges due to its broader range of scenarios, heterogeneous recording devices, and high intra-class variability of violent behaviors. 

\begin{table}[t]
\centering
\begin{threeparttable}
\caption{Performance comparison of SOTA methods on the UCF-Crime dataset.}
\label{tab:ucf_crime}
\begin{tabular}{cccccc}
\toprule
Supervision & Methods & Feature Encoder &  Ground Truth & AUC(\%) & FAR(\%) \\ \midrule
\multirow{2}{*}{Semi} & Object-Centric~\cite{ionescu2019object} & - & N & 61.60 & - \\
 & Conv-AE~\cite{hasan2016learning} & AE & N & 50.60 & 27.2 \\ \midrule
\multirow{9}{*}{Weak} & HL-Net~\cite{wu2020not} & I3D & B & 82.44 & - \\
 & RTFM~\cite{tian2021weakly} & I3D & B & 84.30 & - \\
 & DDL~\cite{pu2022locality} & I3D & B & 85.12 & - \\
 & MGFN~\cite{chen2023mgfn} & I3D & B & 86.98 & - \\
 & S3R~\cite{wu2022self} & I3D & B & 80.26 & - \\
 & UR-DMU~\cite{zhou2023dual} & I3D & B & 86.97 & 1.05 \\
 & CU-Net~\cite{zhang2023exploiting} & I3D & B & 86.22 & - \\
 & SAA~\cite{fan2024weakly} & I3D & B & 86.19 & - \\ 
 & OE-CTST~\cite{majhi2024oe} & I3D & B & 86.37 & - \\
 \midrule
\multirow{1}{*}{Weak} & \textbf{Ours} & I3D & B & 87.24 & 0.54 \\ \bottomrule
\end{tabular}
\begin{tablenotes}[flushleft]
      \footnotesize
      \item \textbf{Notes:} N: Normal; B: Binary.
\end{tablenotes}
\end{threeparttable}
\end{table}

\textbf{Evaluation Metrics.} Consistent with prior work~\cite{pu2022locality,zhou2023dual}, the Area Under the Receiver Operating Characteristic curve (AUC) is adopted as the standard frame-level evaluation metric for the UCF-Crime dataset in WSVAD. For the XD-Violence dataset, Average Precision (AP) is employed as the primary metric, owing to its heightened sensitivity to the precision-recall trade-off and greater suitability for large-scale, multi-scene data. To further assess model robustness in normal scenarios, the False Alarm Rate (FAR) on normal test videos is also reported, defined as the percentage of normal frames misclassified as anomalous. A lower FAR indicates superior stability and reliability.
\subsection{Implementation Details}

Following prior works~\cite{zhou2023dual,tian2021weakly}, we adopt the I3D backbone \cite{carreira2017quo} pre-trained on Kinetics-400 for spatio-temporal feature extraction. 
The TRM employs hidden dimensions of 128 and 256 for UCF-Crime and XD-Violence, respectively, with corresponding attention heads set to 1 and 16. The dropout rate in the MLP component is fixed at 0.1. For the causal convolution classifiers, the kernel sizes are empirically set to 9 on UCF-Crime and 5 on XD-Violence. In the dynamic boundary threshold, the hyperparameter
$\lambda$ is set to 3. The sliding window sizes $W$ for the moving average are set to 5 (UCF-Crime) and 3 (XD-Violence). The exponent hyperparameters $\alpha$ are set to 2 and 1, respectively, while the weighting parameters $(\lambda_s, \lambda_e, \lambda_1, \lambda_2)$ are set to 
$[10,5,0.05,0.05]$ and 
$[20,1,0.01,0.05]$ for the two datasets, correspondingly.
The proposed framework is implemented in PyTorch and optimized with the Adam optimizer. We train the model for 50 epochs with a mini-batch size of 128. The initial learning rate is set to $5 \times 10^{-4}$ and is adjusted throughout the training process using a cosine annealing decay strategy. To balance computational efficiency and detection accuracy, a snippet sampling threshold of 200 is adopted during training, following established practice in \cite{wu2020not,pu2022locality}. All experiments, including training and inference latency evaluations, are conducted on a workstation equipped with an NVIDIA Tesla A40 GPU.

\begin{table}[t]
\centering
\begin{threeparttable}
\caption{Performance comparison of SOTA methods on the XD-Violence dataset.}
\label{tab:xd_violence}
\begin{tabular}{cccccc}
\toprule
Supervision & Methods &  Feature Encoder &  Ground Truth & AP(\%) & FAR(\%) \\ \midrule
\multirow{2}{*}{Semi} & OCSVM~\cite{scholkopf1999support} & I+V & N & 27.25 & - \\
 & Conv-AE~\cite{hasan2016learning} & I+V & N & 30.77 & - \\ \midrule
\multirow{13}{*}{Weak} & HL-Net~\cite{wu2020not} & I3D & B & 75.41 & - \\
 & RTFM~\cite{tian2021weakly} & I3D & B & 77.81 & - \\
 & DDL~\cite{pu2022locality} & I3D & B & 80.72 & - \\
 & MGFN~\cite{chen2023mgfn} & I3D & B & 79.19 & - \\
 & S3R~\cite{wu2022self} & I3D & B & 80.26 & - \\
 & UR-DMU~\cite{zhou2023dual} & I3D & B & 81.66 & 0.65 \\
 & CU-Net~\cite{zhang2023exploiting} & I3D & B & 78.74 & - \\
 & OE-CTST~\cite{majhi2024oe} & I3D & B & 80.56 & - \\
 & SAA~\cite{fan2024weakly} & I3D & B & 83.59 & - \\ \cmidrule{2-6}
 & HL-Net~\cite{wu2020not} & I+V & B & 78.64 & - \\
 & UR-DMU~\cite{zhou2023dual} & I+V & B & 81.77 & - \\
 & CU-Net~\cite{zhang2023exploiting} & I+V & B & 81.43 & - \\
 & SAA~\cite{fan2024weakly} & I+V & B & 83.77 & - \\
 \midrule
\multirow{1}{*}{Weak} & \textbf{Ours} & I3D & B & 83.89 & 0.46 \\ \bottomrule
\end{tabular}
\begin{tablenotes}[flushleft]
      \footnotesize
      \item \textbf{Notes:} I+V: I3D+VGGish; N: Normal; B: Binary.
\end{tablenotes}
\end{threeparttable}
\end{table}

\subsection{Comparison with SOTA Methods}
The performance comparisons between the proposed method and contemporary SOTA approaches are summarized in Tables~\ref{tab:ucf_crime} and \ref{tab:xd_violence}. 
The quantitative results on the UCF-Crime dataset are reported in Table~\ref{tab:ucf_crime}. Under identical experimental settings—utilizing the I3D backbone for feature extraction and relying solely on video-level labels (normal vs. abnormal) as ground truth—our method achieves the best overall performance. Specifically, it attains an AUC of 87.24\%, surpassing the previous SOTA by 0.27\%. Moreover, it achieves an FAR of 0.54\%, which is notably lower than the 1.05\% reported by UR-DMU. These results confirm the effectiveness and robustness of the proposed approach in challenging real-world surveillance scenarios.

For the XD-Violence dataset, the comparative results are represented in Table~\ref{tab:xd_violence}. Under the same experimental setup, our approach consistently outperforms existing SOTA methods, achieving a peak AP of 83.89\% and a best FAR of 0.46\%. This strong performance across both benchmarks further demonstrates the robustness and generalization capability of the proposed framework for large-scale and complex video anomaly detection tasks.

\subsection{Ablation Study}
To validate the effectiveness of individual components in the proposed framework, a series of ablation experiments is conducted on the UCF-Crime dataset, as summarized in Table~\ref{tab:ablation1}. 
Comparing Exp1 and Exp2 highlights the impact of introducing the TRM for explicit long-range temporal modeling. The refined snippet representations substantially enhance the model’s expressive capacity, resulting in a performance improvement of 2.28\% compared to the baseline. A comparison between Exp2 and Exp3 further demonstrates that incorporating a class token as a global semantic prior improves the AUC by 0.74\%. Finally, compared to Exp3, Exp4 introduces the ESM to adaptively capture event-level features. The inclusion of ESM further boosts the anomaly detection performance to 87.24\%, underscoring the importance of multi-granularity feature modeling. Combining all proposed components yields the highest overall detection performance.

\begin{table}[t]
    \centering
    \caption{Ablation study of different modules in our method on the UCF-Crime dataset.} 
    \label{tab:ablation1}
    \begin{tabular}{cccccccc}
    \hline

        No & I3D & TRM & ESM & class token & AUC(\%) & FAR(\%) \\ \hline
        1& \checkmark & - & - &   -  & 82.50 & 1.84\\
        2& \checkmark &\checkmark & - &   - & 84.78 & 0.85 \\
        3& \checkmark &\checkmark &  - & \checkmark & 85.52 & 0.53\\
        4& \checkmark &\checkmark &  \checkmark & \checkmark  & 87.24 & 0.34 \\
        \hline
    \end{tabular}
\end{table}

\begin{figure}[t]
    \centering
    \includegraphics[width=\textwidth]{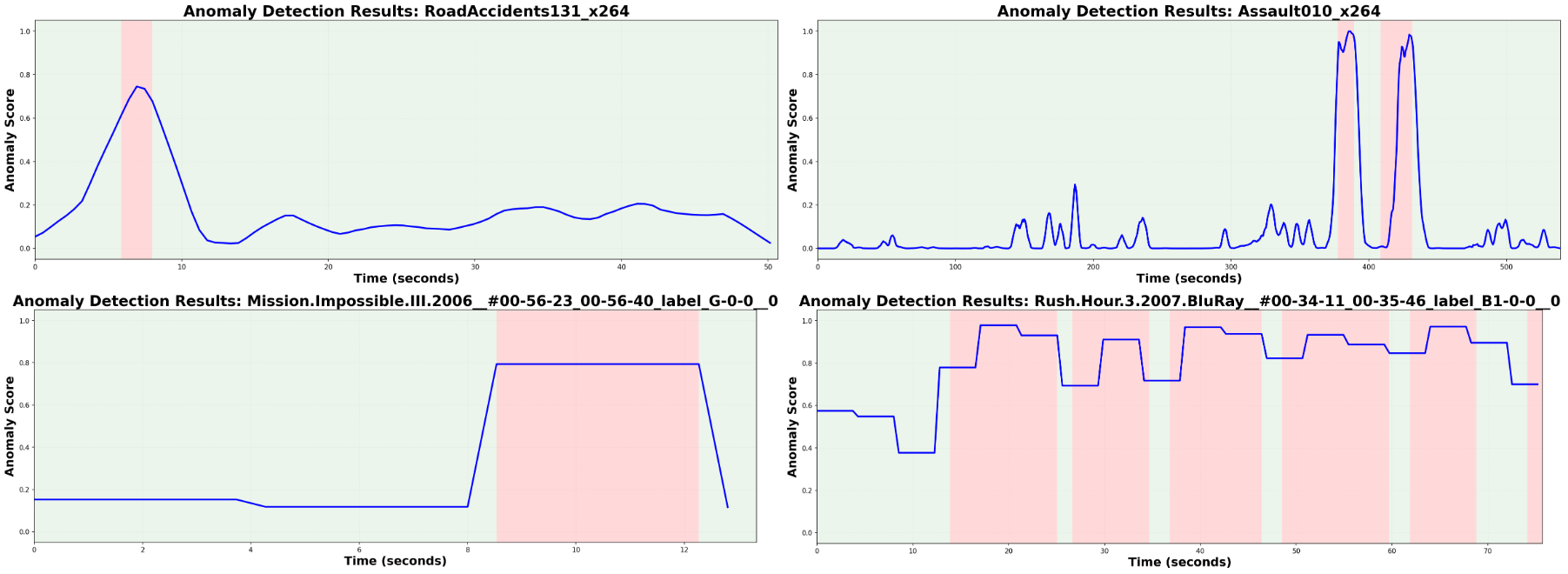}
    \caption{Visualized anomaly detection results of our proposed method. Examples in the first row are from the UCF-Crime dataset, while those in the second row are from the XD-Violence dataset. In each plot, the blue curve indicates the anomaly score at each time step, with green backgrounds representing normal segments and red backgrounds denoting anomalous segments.}
    \label{fig:visual}
\end{figure}

\subsection{Qualitative Analysis} 
To qualitatively evaluate the effectiveness of our proposed framework, detection results are visualized across two benchmark datasets (see Fig.~\ref{fig:visual}). The visualization demonstrates that our method robustly identifies anomalous events and precisely localizes their temporal intervals, regardless of whether the video contains a single anomaly or multiple anomalous events. This high degree of alignment between the predicted anomaly scores and the ground truth intervals further validates the superior discriminative capability and generalization of our multi-scale model in complex real-world scenarios.

\section{conclusion}

In this paper, we presented an adaptive anomaly detection framework for WSVAD. Addressing the limitations of conventional snippet-level analysis and fixed aggregation heuristics, we introduced the TRM and ESM. The TRM effectively models long-range temporal dependencies while distilling a holistic video-level representation through a learnable class token and dynamic position encoding. Complementary to this, the ESM adaptively identifies event boundaries to capture event-level features, accommodating the inherent variability in anomaly frequency and duration. Furthermore, we replaced the traditional \textit{top-k} pooling with an adaptive similarity-based fusion strategy, enabling more robust and generalized video-level predictions. Extensive experiments demonstrate the performance of our model and the effectiveness of each component. For future work, we plan to explore the integration of multimodal signals, such as audio and optical flow, to further enhance the robustness of anomaly detection in complex, cluttered environments.

\bibliographystyle{splncs04}
\bibliography{refs}
\end{document}